\PassOptionsToPackage{unicode}{hyperref}
\PassOptionsToPackage{hyphens}{url}
\PassOptionsToPackage{dvipsnames,svgnames,x11names}{xcolor}
\documentclass[
  11pt,
]{article}
\usepackage{xcolor}
\usepackage[margin=1in]{geometry}
\usepackage{amsmath,amssymb}
\usepackage{iftex}
\ifPDFTeX
  \usepackage[T1]{fontenc}
  \usepackage[utf8]{inputenc}
  \usepackage{textcomp} 
\else 
  \usepackage{unicode-math} 
  \defaultfontfeatures{Scale=MatchLowercase}
  \defaultfontfeatures[\rmfamily]{Ligatures=TeX,Scale=1}
\fi
\usepackage{lmodern}
\ifPDFTeX\else
\fi
\IfFileExists{upquote.sty}{\usepackage{upquote}}{}
\IfFileExists{microtype.sty}{
  \usepackage[]{microtype}
  \UseMicrotypeSet[protrusion]{basicmath} 
}{}
\usepackage{setspace}
\makeatletter
\@ifundefined{KOMAClassName}{
  \IfFileExists{parskip.sty}{%
    \usepackage{parskip}
  }{
    \setlength{\parindent}{0pt}
    \setlength{\parskip}{6pt plus 2pt minus 1pt}}
}{
  \KOMAoptions{parskip=half}}
\makeatother
\makeatletter
\ifx\paragraph\undefined\else
  \let\oldparagraph\paragraph
  \renewcommand{\paragraph}{
    \@ifstar
      \xxxParagraphStar
      \xxxParagraphNoStar
  }
  \newcommand{\xxxParagraphStar}[1]{\oldparagraph*{#1}\mbox{}}
  \newcommand{\xxxParagraphNoStar}[1]{\oldparagraph{#1}\mbox{}}
\fi
\ifx\subparagraph\undefined\else
  \let\oldsubparagraph\subparagraph
  \renewcommand{\subparagraph}{
    \@ifstar
      \xxxSubParagraphStar
      \xxxSubParagraphNoStar
  }
  \newcommand{\xxxSubParagraphStar}[1]{\oldsubparagraph*{#1}\mbox{}}
  \newcommand{\xxxSubParagraphNoStar}[1]{\oldsubparagraph{#1}\mbox{}}
\fi
\makeatother

\usepackage{longtable,booktabs,array}
\usepackage{calc} 
\usepackage{etoolbox}
\makeatletter
\patchcmd\longtable{\par}{\if@noskipsec\mbox{}\fi\par}{}{}
\makeatother
\IfFileExists{footnotehyper.sty}{\usepackage{footnotehyper}}{\usepackage{footnote}}
\makesavenoteenv{longtable}
\usepackage{graphicx}
\makeatletter
\newsavebox\pandoc@box
\newcommand*\pandocbounded[1]{
  \sbox\pandoc@box{#1}%
  \Gscale@div\@tempa{\textheight}{\dimexpr\ht\pandoc@box+\dp\pandoc@box\relax}%
  \Gscale@div\@tempb{\linewidth}{\wd\pandoc@box}%
  \ifdim\@tempb\p@<\@tempa\p@\let\@tempa\@tempb\fi
  \ifdim\@tempa\p@<\p@\scalebox{\@tempa}{\usebox\pandoc@box}%
  \else\usebox{\pandoc@box}%
  \fi%
}
\def\fps@figure{htbp}
\makeatother

\NewDocumentCommand\citeproctext{}{}

\makeatletter
 \let\@cite@ofmt\@firstofone
 \def\@biblabel#1{}
 \def\@cite#1#2{{#1\if@tempswa , #2\fi}}
\makeatother
\newlength{\cslhangindent}
\newlength{\csllabelwidth}
\newenvironment{CSLReferences}[2] 
 {\begin{list}{}{%
  \setlength{\itemindent}{0pt}
  \setlength{\leftmargin}{0pt}
  \setlength{\parsep}{0pt}
  \ifodd #1
   \setlength{\leftmargin}{\cslhangindent}
   \setlength{\itemindent}{-1\cslhangindent}
  \fi
  \setlength{\itemsep}{#2\baselineskip}}}
 {\end{list}}
\usepackage{calc}

\newcommand{\CSLLeftMargin}[1]{\parbox[t]{\csllabelwidth}{\strut#1\strut}}
\newcommand{\CSLRightInline}[1]{\parbox[t]{\linewidth - \csllabelwidth}{\strut#1\strut}}

\providecommand{\tightlist}{%
  \setlength{\itemsep}{0pt}\setlength{\parskip}{0pt}}

\usepackage{booktabs}
\usepackage{longtable}
\usepackage{array}
\usepackage{multirow}
\usepackage{wrapfig}
\usepackage{float}
\usepackage{colortbl}
\usepackage{pdflscape}
\usepackage{tabu}
\usepackage{threeparttable}
\usepackage{threeparttablex}
\usepackage[normalem]{ulem}
\usepackage{makecell}
\usepackage{xcolor}
\usepackage{pdflscape}
\usepackage{float}
\makeatletter
\@ifpackageloaded{caption}{}{\usepackage{caption}}
\AtBeginDocument{%
\ifdefined\contentsname
  \renewcommand*\contentsname{Table of contents}
\else
  \newcommand\contentsname{Table of contents}
\fi
\ifdefined\listfigurename
  \renewcommand*\listfigurename{List of Figures}
\else
  \newcommand\listfigurename{List of Figures}
\fi
\ifdefined\listtablename
  \renewcommand*\listtablename{List of Tables}
\else
  \newcommand\listtablename{List of Tables}
\fi
\ifdefined\figurename
  \renewcommand*\figurename{Figure}
\else
  \newcommand\figurename{Figure}
\fi
\ifdefined\tablename
  \renewcommand*\tablename{Table}
\else
  \newcommand\tablename{Table}
\fi
}
\@ifpackageloaded{float}{}{\usepackage{float}}
\floatstyle{ruled}
\@ifundefined{c@chapter}{\newfloat{codelisting}{h}{lop}}{\newfloat{codelisting}{h}{lop}[chapter]}
\floatname{codelisting}{Listing}

\makeatother
\makeatletter
\@ifpackageloaded{caption}{}{\usepackage{caption}}
\@ifpackageloaded{subcaption}{}{\usepackage{subcaption}}
\makeatother
\usepackage{bookmark}
\IfFileExists{xurl.sty}{\usepackage{xurl}}{} 
\hypersetup{
  pdftitle={Children, but not language models, show accelerating returns in word learning},
  pdfauthor={Michael C. Frank},
  pdfkeywords={Early word learning; Bayesian data analysis; Large
language models},
  colorlinks=true,
  linkcolor={blue},
  filecolor={Maroon},
  citecolor={Blue},
  urlcolor={Blue},
  pdfcreator={LaTeX via pandoc}}

\title{Children, but not language models, show accelerating returns in
word learning}
\author{Michael C. Frank}
\date{}
\begin{document}
\maketitle
\begin{abstract}
\noindent Children learn hundreds of words over the first years of their
lives, in a process that begins slowly but quickly picks up speed. Prior
models describe vocabulary growth as evidence accumulation over time.
Here we show that the process is best characterized as
\emph{accelerating} accumulation: children learn more from each
additional unit of linguistic experience than they did from the one
before. In contrast to children, language models -- even those trained
on child-directed speech -- do not accelerate. Instead, they show
constant proportional returns on new data, consistent with scaling laws.
Children learn using many orders of magnitude less training data than
language models; their increasingly efficient use of their learning
input is a candidate explanation.
\end{abstract}

\setstretch{1.05}
Children's language grows rapidly during early childhood (\emph{1},
\emph{2}), and the ``explosion'' of expressive vocabulary is a signature
human cognitive achievement (\emph{3}, \emph{4}). Providing a
quantitative characterization of this process is an important scientific
goal (\emph{5}--\emph{9}), but until recently, language acquisition
could only be studied in human learners. Language models (LMs) show
increasing linguistic sophistication, however, allowing them to become a
model system for the study of human language (\emph{10}, \emph{11}).

The most successful LMs differ from human learners in an important way:
they are trained on vastly more data than human learners receive during
childhood. While even open-source models are routinely trained on
trillions of words of data (\emph{12}), children hear on the order of
100k -- 1M words of input per month, with 30-40M words as an approximate
upper bound by age three (\emph{13}, \emph{14}). Why do models require
so many more orders of magnitude of training data to achieve basic
linguistic competence?

Here we study this ``data gap'' by comparing word learning as a function
of training data in both children and LMs. Vocabulary is a good target
for comparison because the total number of different words a child
produces is an important measure of expressive abilities that is also
tightly coupled to grammatical development and other aspects of early
language (\emph{1}, \emph{15}). Vocabulary size varies widely between
children of the same age (\emph{1}), and relative vocabulary size is a
predictor of later academic success (\emph{16}, \emph{17}). Large
longitudinal vocabulary datasets using the MacArthur-Bates Communicative
Development Inventories (CDIs), a popular parent-report form, make it
possible to characterize this growth even in young children (\emph{1},
\emph{18}, \emph{19}).

We compare how vocabulary grows with input data in children and LMs. We
begin by developing a psychometric model of vocabulary growth in
children (an ``accelerating accumulator''), which shows a good fit to
population variation across children. This model suggests that
children's accumulation of evidence for words, not just their observed
vocabulary size, accelerates across early childhood. This acceleration
leads to an increasing ``return'' on learning from linguistic
experience: the same experience counts for more later in development. In
contrast to children, LMs do not accelerate: they show constant
proportional returns on additional data. This analysis exposes a
critical disanalogy between LMs and children and suggests new avenues
for the study of both.

\section{An accelerating accumulator model of vocabulary
growth}\label{an-accelerating-accumulator-model-of-vocabulary-growth}

Recent models describe children's vocabulary growth as a process of
accumulation (\emph{8}, \emph{9}, \emph{20}, \emph{21}). Each distinct
word (type) can be conceptualized as a ``bucket,'' and every token of
language input then is a ``drop'' that falls into its respective bucket.
When each bucket is full, the word is learned (\emph{9}, \emph{20}).
Accumulator models have been used to recover interpretable differences
in ``bucket size,'' often based on word properties including
concreteness and syntactic category (\emph{8}, \emph{9}). They also
provide a graded explanation for the widely observed ``vocabulary
spurt'' (\emph{20}, \emph{22}, \emph{23}). Children's \emph{observed
vocabulary} (in terms of count) grows supralinearly with age (\emph{1},
\emph{18}), but this qualitative pattern is consistent with many
possible underlying mechanisms (\emph{20}).

We begin an accumulator model that operates over ratio-scale variables
with real units: words (tokens heard) per hour and total words (types)
in the child's vocabulary (\emph{9}). The presence of a true zero in
each of these variables means that they can be log transformed, which is
critical for capturing the shape of their growth over development
(\emph{24}, \emph{25}). This accumulator can be written as a Rasch model
from psychometrics (\emph{26}), which assumes that each child \(i\) has
a learning ability \(\theta_i\) and each word \(j\) has a difficulty
\(\delta_j\).\footnote{Prior work has focused on estimating properties
  of individual words (\emph{8}, \emph{9}, \emph{27}), revealing that
  there is predictable word-level difficulty information but that only a
  modest portion of the variation between words is carried by frequency
  information. Words can be hard for a child to acquire because they
  vary also in the complexity and concreteness of their meanings, their
  syntactic category and syntactic context, their phonology, and many
  other factors. In the current work, we estimate word difficulty as a
  free parameter, and focus instead on variation between children.} The
probability that child \(i\) produces word \(j\) is:

\[P(w_{i,j}=1 \mid \theta_i, \delta_j) =\frac{\exp(\theta_i - \delta_j)}{1 + \exp(\theta_i - \delta_j)}.\]

We then describe the child's ability as

\[\theta_i(t) = \xi_i + \kappa_i ~ \log(t/a_0) + \log(H),\]

\noindent which contains three terms: an intercept \(\xi_i\) describing
the child's baseline learning efficiency, an acceleration term,
\(\kappa_i ~ \log(t/a_0)\) that describes the rate of accumulation
across developmental time (\(t\), relative to an anchor \(a_0\)), and a
constant \(H\) for the number of waking hours over which accumulation
occurs.

This model is an ``accelerating accumulator'' model, in which
accumulating language ``counts for more'' later in development;
\(\kappa\) is the key exponent describing developmental scaling. In
contrast, an accumulator with \(\kappa=1\) would be
\(\theta_i(t) = \xi + log(t/a_0) + log(H)\). This ``pure accumulator''
yields linear returns on experience, while \(\kappa>1\) shows
acceleration. Under \(\kappa=1\), the size of the ``drops'' filling the
buckets remains constant across development; if \(\kappa>1\), later
``drops'' are bigger.

We posit that children vary on both their efficiency of accumulation
\(\xi_i\) and their acceleration \(\kappa_i\), creating a variant of a
standard longitudinal growth model in which children vary on their slope
and intercept (\emph{28}) (Figure~\ref{fig-schematic} A). Our model is
close to the formulations of (\emph{8}) and (\emph{9}), but has not yet
been proposed (supplementary text).

\section{Fitting the accumulator
model}\label{fitting-the-accumulator-model}

CDIs are a widely used, reliable, and valid method for taking an
in-depth snapshot of children's early vocabulary (\emph{1}, \emph{18},
\emph{19}). Our analysis depends on the availability of longitudinal CDI
data at the individual word level. Summed vocabulary scores do not allow
for the estimation of word difficulties \(\delta\) and cross-sectional
data do not allow for model identification (fig.~S1). Thus, we use
item-level longitudinal data from children with three or more
longitudinal administrations. We analyze reported production, which is
both measured over a wider age range and tends to be a more reliable
measure than reported comprehension (\emph{1}).

We use five longitudinal datasets of monolingual, typically developing
children ages 8--36 months (three from American English and one each
from Norwegian and Japanese, \(N_{total} = 1841\); see fig.~S2 and table
S1), downloaded from Wordbank (\emph{29}). We fit the accelerating
accumulator model (M3) to each dataset using Bayesian inference. To test
whether each model component was necessary, we additionally fit three
model ablations, removing individual variation in acceleration (without
\(\kappa_i\); M2), removing individual variation in efficiency (without
\(\xi_i\); M1), and finally removing acceleration entirely (without
\(\kappa\); M0).

\begin{figure}[H]

\centering{

\includegraphics[width=1\linewidth,height=\textheight,keepaspectratio]{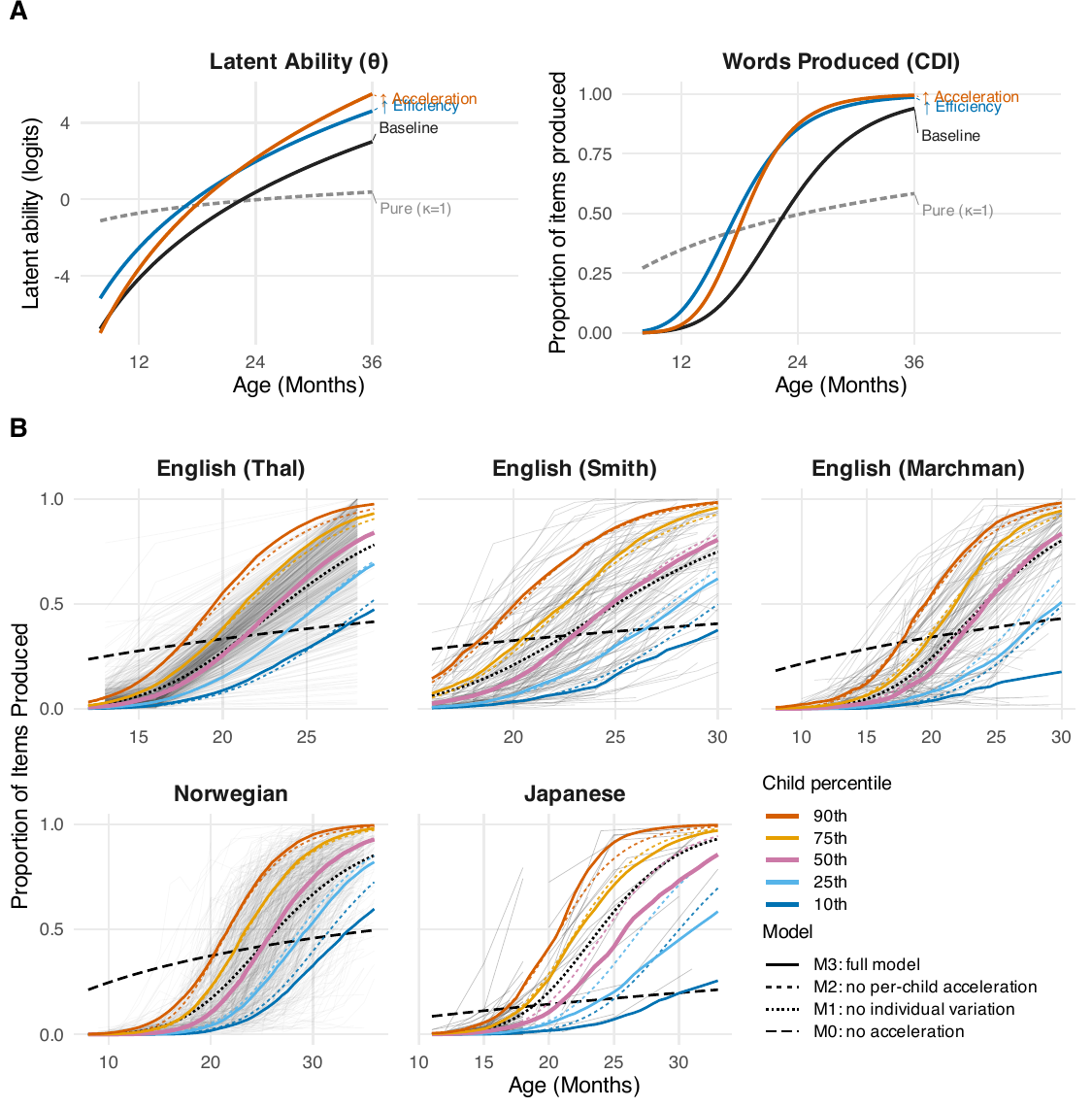}

}

\caption{\label{fig-schematic}\textbf{The accelerating accumulator
provides a psychometric model of vocabulary growth, describing the
children's word learning across datasets and languages.} (A) Model
predictions in latent-ability (\(\theta\)) space and, projected through
the logistic item response model, in words-produced (CDI) space. Both
pure accumulator (\(\kappa = 1\), dashed) and accelerating accumulator
predictions are plotted. Greater efficiency (\(\xi\)) lifts the curve's
level, while greater acceleration (\(\kappa\)) fans it open. (B)
Individual children's observed trajectories (gray) for each of the five
longitudinal datasets, with quantiles computed across the per-child
parameters of the fitted models (colored percentile bands for M2 and M3;
black lines for M0 and M1, which do not vary across individuals).}

\end{figure}

The accelerating accumulator provided the best fit to children's growth
trajectories (Figure~\ref{fig-schematic} B, see table S2), with
substantial improvements in item-level expected log predictive density
between the accelerating accumulator and the next best model (LOO ELPD
\(>760\) for all five datasets). Indeed, fits from M3 corresponded
closely to curves derived from a non-parametric GAMLSS beta regression,
a flexible model class used to produce percentile norms for the CDI
instruments (\emph{19}) (fig.~S3).

Estimates of acceleration were high for all datasets, robustly rejecting
the pure accumulator (\(\kappa=1\)) in every case: 10.6 {[}10.5, 10.8{]}
-- 13.3 {[}13.2, 13.3{]}. \(\kappa\) estimates were essentially
unchanged using a larger dataset and in a hierarchical model pooling
across all datasets (table S5). A 2PL IRT model achieved a better fit
but \(\kappa\) estimates remained far above the pure accumulator and
word-level age of acquisition estimates were very similar (table S8);
the specifics of item selection on CDI forms had only minimal effects on
\(\kappa\) estimates (fig.~S4). Linear (rather than logarithmic) growth
models did not provide better fit (table S7).

Variability is a ubiquitous feature of early language (\emph{1},
\emph{2}). The between-child standard deviation in acceleration
(\(\sigma_b\)) ranged from 3.2 to 6.6 across the five samples; the
fitted population distribution placed 99.5\% of children above the
pure-accumulator value of \(\kappa = 1\). Although the population fits
showed substantial slope variation, typical CDI series are too sparse to
predict individual children's acceleration reliably beyond the
population mean (fig.~S5 and tables S3, S4 and S9). Nevertheless, the
addition of a population acceleration parameter improved individual
trajectory prediction in 91.9\% of children when five or more prior CDIs
were available.

\section{Accelerating accumulation implies less exposure is needed for
later-acquired
words}\label{accelerating-accumulation-implies-less-exposure-is-needed-for-later-acquired-words}

\begin{figure}[H]

\centering{

\includegraphics[width=1\linewidth,height=\textheight,keepaspectratio]{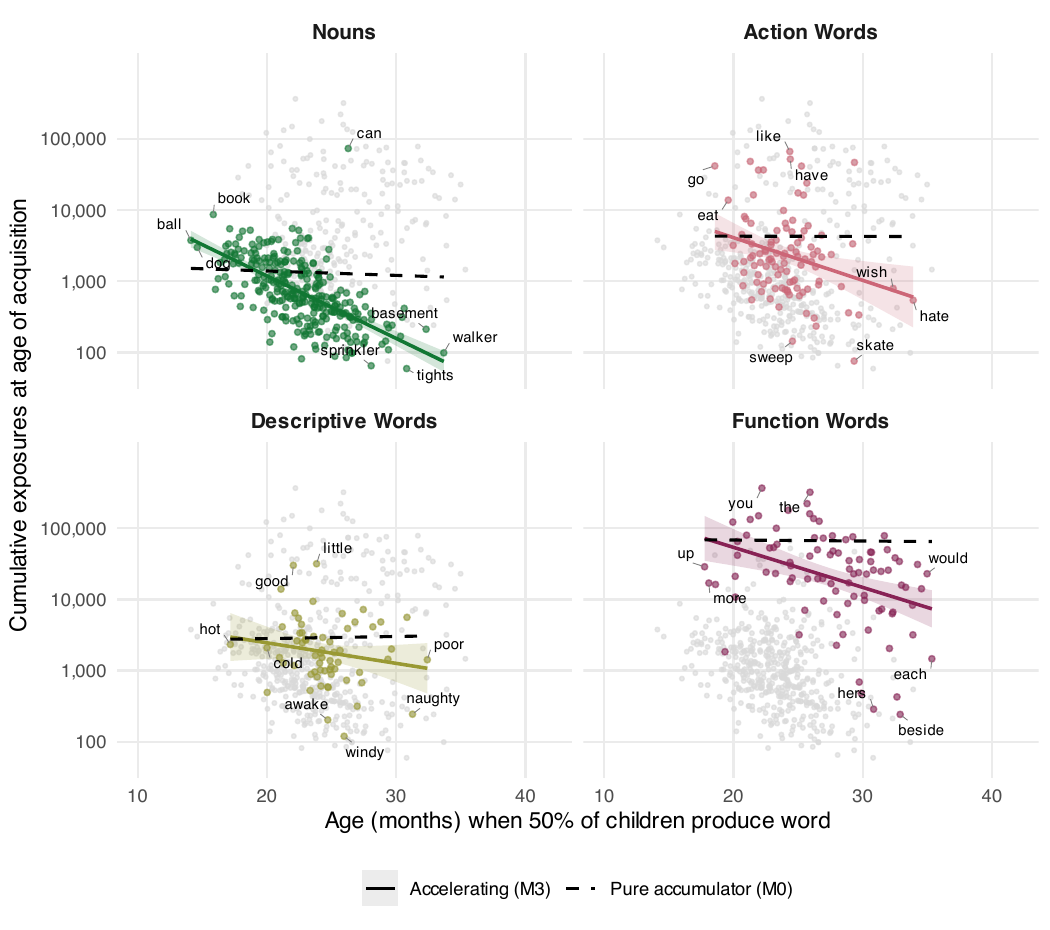}

}

\caption{\label{fig-efficiency}\textbf{Under the accelerating
accumulator model, the number of exposures needed to learn a word
decreases with age.} Plots show estimated cumulative exposures to a
word, plotted by the age at which 50\% of the children in our three
English-language datasets were estimated to produce the word. Subplots
show different word classes, and solid and dashed lines show simple
linear fits from the accelerating (M3) and pure accumulator (M0) models,
respectively; gray dots show the full distribution of words across all
word classes.}

\end{figure}%

The accelerating accumulator model captures a salient fact about
language acquisition: the number of exposures necessary to learn a word
decreases dramatically over childhood. A toddler may have heard a word
like ``book'' thousands of times before they say it; a few years later a
preschooler may hear ``ibex'' a handful of times on a trip to the zoo
and repeat it the next day.

To quantify this phenomenon, we computed each word's age of acquisition
(AoA) \(A\) from our fitted models, with AoA defined as the point at
which 50\% of children produce the word -- in other words, the age at
which child ability and word difficulty balance:
\(A(j) = a_0 \exp\!\big((\delta_j - \log H - \mu_\xi)/\kappa\big)\).
Then for each word, we estimated the cumulative word frequency for an
individual child based on transcripts of speech to children (\emph{30}).

Figure~\ref{fig-efficiency} shows the relationship between cumulative
exposures and age of acquisition for the accelerating accumulator.
Within each word class (e.g., nouns, verbs, and function words), the
number of examples experienced before learning drops exponentially with
development.\footnote{Indeed, this analysis likely understates the
  degree of change in efficiency as it assumes that each exposure to a
  word is equally meaningful. In fact, individual exposures vary widely
  in how much information they give. Sometimes words are overheard
  rather than directed to the child, or lack physical context to allow
  for clear inference (\emph{21}, \emph{31}).} In contrast, the fitted
pure accumulator predicts a far shallower decline; it cannot reproduce
the observed range of acquisition ages, predicting that many early
acquired words will not be learned until adolescence (fig.~S7 and table
S10).

\section{Language models do not
accelerate}\label{language-models-do-not-accelerate}

\begin{figure}[H]

\centering{

\includegraphics[width=1\linewidth,height=\textheight,keepaspectratio]{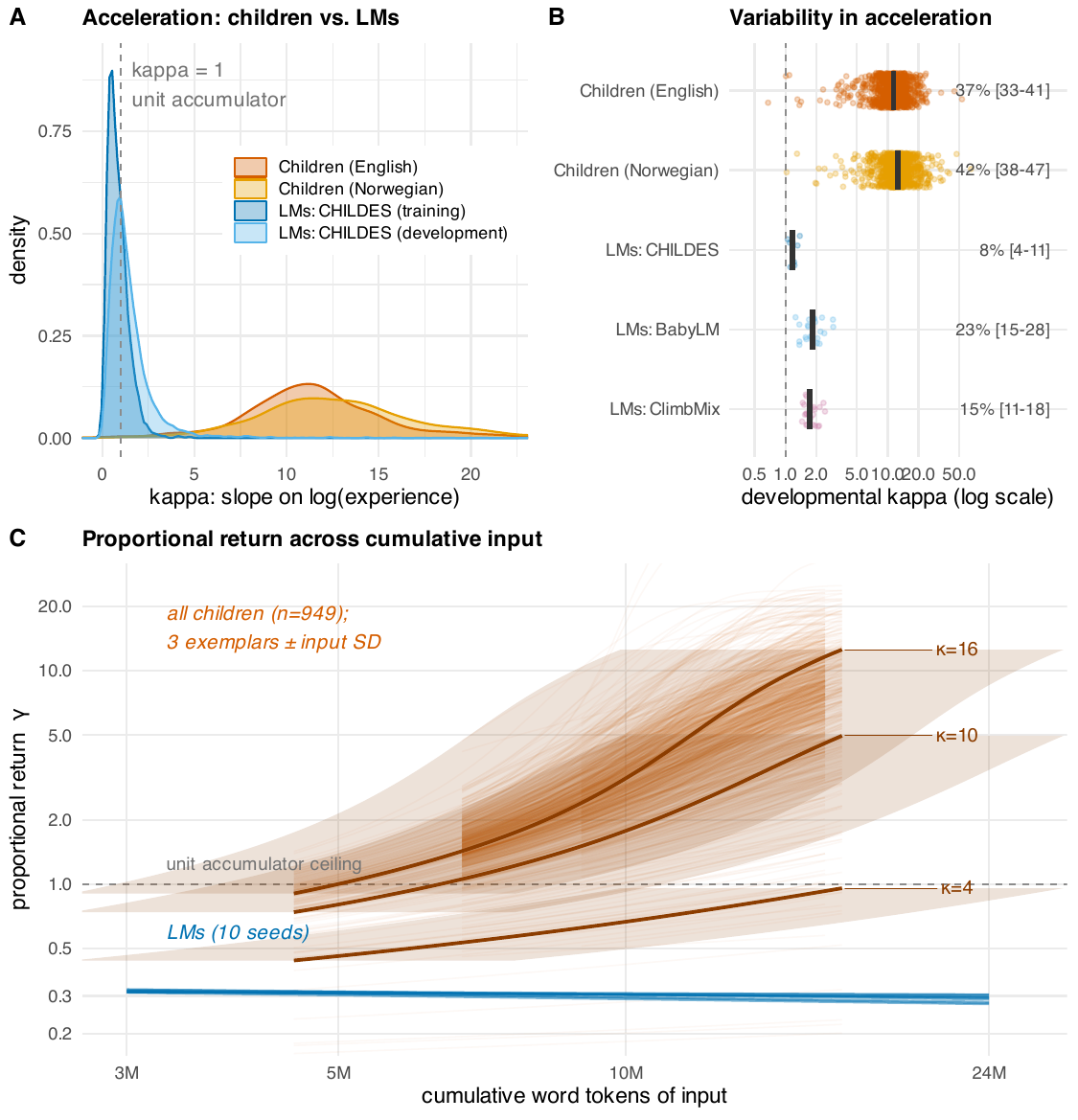}

}

\caption{\label{fig-llm-acceleration}\textbf{Language models (LMs) show
no acceleration in their vocabulary growth.} (A) Density of the
acceleration exponent \(\kappa\) for both children (per child, English
and Norwegian) and LMs (by word). (B) Variation in acceleration (kappa)
across children and LMs. Numbers show coefficients of variation and 95\%
CIs. (C) Proportional return on training (the ratio of decrease in loss
to excess loss) for both children and LMs. Blue lines show LM
trajectories; red lines show children's fitted model trajectories from
M3. Three children are highlighted in dark red with horizontal
uncertainty bands indicating +/- 1 SD in calibration of their input.
Logistic acquisition alone can produce some increase in proportional
return due to ceiling effects; these asymptote at 1 (gray dashed line).}

\end{figure}%

Do LMs show the same signatures of acceleration and variability as
children? To evaluate this question, we begin by building a bridge
between well-known ``scaling laws'' for large language models
(\emph{32}) and the model we describe here for vocabulary growth. The
loss \(L\) of a neural network language model on a dataset size \(D\) is
\(L = E + \frac{B}{D^\beta}\), where \(B\) and \(\beta\) are the free
parameters of a power law and \(E\) is the entropy of the text being
modeled (\emph{32}). This power law scaling behavior can emerge from a
set of independently learned ``quanta'' that are themselves power-law
distributed (\emph{33}); the words of natural language follow such a
power law and are potential quanta of this type.

To trace the emergence of individual words in LMs, we fit sigmoidal
curves to the trajectory of surprisal values (negative log probability
of individual words) across training (following \emph{34}). These
sigmoids provide a link to the logistic curves in the Rasch model we
used with children. The slope of the sigmoid for each word has the same
units as our acceleration parameter \(\kappa\) -- it captures how fast
evidence for that particular word accumulates with training data.

We computed the distribution of these slopes on GPT-2 models (\emph{35})
trained on 24M words of English child-directed speech from the CHILDES
archive (\emph{10}, \emph{30}, \emph{36}), approximating a
three-year-old child's total language input (fig.~S9). Per-word
\(\kappa\) estimates computed across training checkpoints clustered
around 1 (see Figure~\ref{fig-llm-acceleration} A, ``training''
sequence): knowledge about individual words accumulated gradually. This
analysis computes learning over multiple training passes through the
same dataset, however. We thus trained ``developmental'' sequences of
GPT-2 models. Each model in a sequence was trained to convergence on an
increasingly larger subsample of the CHILDES data, ranging from .5M
words to 24M words. \(\kappa\) values computed across these runs also
showed no evidence of acceleration. In contrast, when we applied the
same by-word estimator to data from children, we found high \(\kappa\)
values (fig.~S8 and table S11). \(\kappa\) variation was far higher in
children than LMs, even when training data were varied
(Figure~\ref{fig-llm-acceleration} B).

To make a holistic comparison between the growth of individual models
(rather than words) and individual children, we computed learners'
return on new data. Scaling law analyses describe the change in a
model's loss \(L\) with respect to the fundamental entropy \(E\) of the
training data. These can be manipulated to compute a \emph{proportional
return}, describing the total amount of remaining possible loss that has
been removed by a particular amount of data. To compute this quantity in
children, we computed each child's ``loss'' across all words in the
vocabulary, which we aligned to children's estimated input rate over
time (fig.~S6). On this analysis, LMs had constant proportional returns,
while individual children's proportional returns increased
(Figure~\ref{fig-llm-acceleration} C).

\section{Discussion}\label{discussion}

Children's vocabulary growth over early childhood can be described as a
process of accelerating accumulation. Knowledge of individual words
accumulates over time, but later in development, fewer exposures are
needed to acquire a new word. In contrast, LMs -- even trained on data
from children -- show no such acceleration. Converted to the same
metric, children get increasing returns from additional experience,
whereas conventional LMs show constant proportional returns.

What explains the acceleration in children's word learning? During the
first three years, children's learning, memory, and speed of processing
all change substantially (\emph{24}, \emph{37}--\emph{39}). In the
language of our model, these changes would mean that older children's
``buckets'' require fewer drops to fill them. Changes in the speed and
accuracy of word recognition, in particular, are deeply related to
children's growing vocabulary; faster processing speed allows children
to extract more signal from each incoming utterance (\emph{24},
\emph{39}). All of these changes could lead to some degree of
acceleration.

A second source of acceleration comes from children ``learning to
learn'': using the language they know to become more effective learners.
Their increasingly sophisticated inferential abilities also allow them
to extract progressively more information from the linguistic signal
(\emph{40}, \emph{41}). They learn generalizations that guide their
inferences based on regularities in how their language works
(\emph{42}). They also use the language they already know in the moment
to extract more information from new utterances -- leveraging word
meanings (\emph{40}), syntactic knowledge (\emph{43}), and pragmatic
inference (\emph{41}) to make sophisticated inferences about gaps in
their knowledge (\emph{3}, \emph{5}).

Simple changes in the quantity or ordering of linguistic input are
likely not responsible for acceleration. First, parents do not typically
``fine-tune'' their language to children's developmental level
(\emph{44}, \emph{45}). Second, curricularization of children's language
input, either by age or by other heuristics for simplicity fails to
increase LM performance (\emph{36}, \emph{46}). Still, broader changes
in the richness of input or child-caregiver interaction could facilitate
acceleration; these possibilities should be explored.

In contrast to children, LMs do not mature developmentally, and it is
unclear if standard pre-training provides an analogue to children's
growing inferential abilities. While in-context learning abilities
emerge, they typically do so only with vastly more data than children
receive (\emph{47}, \emph{48}). And few-shot rapid word learning may
emerge LMs via meta-learning, rather than pre-training (\emph{49}).
Establishing linking hypotheses for comparison and alignment of children
and LMs is thus a critical goal for further work (\emph{50}).

The accelerating accumulator is a parsimonious description of variation
in children's vocabulary growth and it accords with the general
structure of the learning problem faced by children. But our evidence is
based on statistical fit and model comparison, rather than a causal
manipulation. More work is required to test the accumulator as a
mechanistic account. Further, here we relied on parent report data from
the CDI. While CDIs have been extensively validated as a measure of
language ability (\emph{1}, \emph{18}, \emph{19}), they include
variation due to changes in overall verbal production ability as well as
vocabulary size. Linking our model to direct measures of vocabulary is
an important next step.

In sum, our study identifies acceleration in learning from experience as
a fundamental difference between children and LMs. Characterizing the
nature of children's accelerating returns and understanding the
circumstances under which it can arise in artificial systems should thus
be a focus for future investigation.

\section{Acknowledgments}\label{acknowledgments}

\textbf{Funding}: Computing for this project was performed on the
Sherlock and Marlowe clusters. We would like to thank Stanford
University and Stanford Research Computing for providing computational
resources and support that contributed to these research results. Funds
for computational experiments were also provided by a gift from Meta to
support research on children's home language environment. \textbf{Author
contributions}: M.C.F designed and performed all research and wrote the
paper. Claude Code was used for code generation, visualization, and
simulation infrastructure. \textbf{Competing interests}: The author
declares no competing interests. \textbf{Data, code, and materials
availability}: Reproducible code for this paper is available at
\url{https://github.com/mcfrank/acceleration}. All model fits are
archived at
\url{https://redivis.com/datasets/datapages.acceleration:a1c7}. Data
were retrieved from Wordbank using the \texttt{wordbankr} package
(\emph{29}) and CHILDES using the~\texttt{childesr} package (\emph{51}).
All trained LMs are available at
\url{https://huggingface.co/mcxfrank/childes-gpt2-ladder} and
\url{https://huggingface.co/mcxfrank/gpt2-composition-control}.

\section{List of Supplementary
Materials}\label{list-of-supplementary-materials}

Materials and Methods

Supplementary Text

Figs. S1 to S9

Tables S1 to S11

\section{References}\label{references}

\phantomsection\label{refs}

\newpage
\begin{center}
{\LARGE\bfseries Supplemental Information}
\end{center}

\section{Materials and Methods}\label{materials-and-methods}

\subsection{Vocabulary data}\label{vocabulary-data}

All longitudinal datasets with three or more observations per child were
included. English (Thal) data are from (\emph{52}), Norwegian is from
(\emph{53}), and Japanese is from (\emph{54}); English (Marchman) and
English (Smith) are unpublished datasets available through Wordbank
(\emph{29}). We applied a quality filter to exclude administrations that
showed either decreases of more than 25\% of productive vocabulary or
increases that exceeded 40\% of total vocabulary per month. This
excluded 36 children in total (1.9\%), nearly all from the English
(Marchman) dataset (16\%; Norwegian 0.5\%, none from the others) (SI:
Data Exclusions).

\subsection{Bayesian models}\label{bayesian-models}

Using Stan (\emph{55}), Bayesian models were fit separately to each
dataset per the specification described in the text. The model was
parameterized to capture individual children's variation from the
population mean such that \(\xi_i = \mu_\xi + a_i\) and
\(\kappa_i = \mu_\kappa + b_i\), where
\((a_i, b_i) \sim \mathrm{MVN}(\mathbf 0, \Sigma)\). To facilitate
interpretation of the accumulator mechanism, we included the constant
\(\log H = \log(365) \approx 5.90\) (waking hours/month) and the anchor
age \(a_0 = 18\) months (so \(\log(t/a_0)=0\) at 18 months).

We chose weakly informative priors throughout, including on the
efficiency and acceleration parameters
\(\mu_\xi \sim \mathcal N(-6,\ 5)\) and
\(\mu_\kappa \sim 1 +  \mathcal N(0,\ 5)\) (centered around the pure
accumulator model) as well as the individual child standard deviations
\(\sigma_a \sim \mathrm{Half\text{-}Normal}(0,\ 3)\) and
\(\sigma_b \sim \mathrm{Half\text{-}Normal}(0,\ 5)\) and the item
difficulties \(\tau_\delta \sim \mathrm{Half\text{-}Normal}(0,\ 5)\). We
also chose a prior that mildly regularized the correlations between
\(a_i\) and \(b_i\) (\(\mathbf L \sim \mathrm{LKJ}(2)\)).

All models were fit using cmdstanr, with four independent chains of
1,000 warmup and 1,000 sampling iterations each (4,000 post-warmup
draws) and a target acceptance rate of 0.9. The primary series of models
was additionally refit using 2,000 warmup and 8,000 sampling iterations.
We assessed convergence using rank-normalized split-\(\hat{R}\) and bulk
effective sample size. For parameters we interpret (primarily
\(\kappa\), \(\sigma_a\), \(\sigma_b\)), we required \(\hat{R} < 1.01\)
and bulk ESS above 400.

\subsection{Learning efficiency}\label{learning-efficiency}

Age of acquisition for each word was computed based on the weighted
average of word difficulties \(\delta_j\) from the three fitted M3
models for English. For the population level of \(\xi\) and \(\kappa\),
we computed the earliest age at which there was at least a .5
probability of producing the word. Average exposures per month to each
word were computed as \(N_j = R \cdot t_{50}(j) \cdot p_j\), where \(R\)
is the rate of input (words/month), \(t_{50}(j)\) is the word-level age
of acquisition computed above, and \(p_j\) is the word's corpus
probability. We used average input estimates derived from (\emph{56})
and (\emph{57}) for our estimate of \(R\) (see SI: Input Estimation),
and computed \(p_j\) from empirical word frequencies in all North
American English corpora in childes-db release 2021.1.

\subsection{Language model training}\label{language-model-training}

Following (\emph{36}), we fit randomly-initialized \texttt{GPT-2-small}
(124M parameters) models, using the 24M-word (\textasciitilde47M-token)
preparation of English-language CHILDES data from that study. Consistent
with the previous work, we included children's own utterances in the
training data. Children do in fact hear their own speech; more
importantly, removing one side of every interaction would create
disjoint, unnatural training data. We also fit models to the 2nd BabyLM
challenge dataset (downloaded from \url{osf.io/ad7qg}, with CHILDES data
removed) (\emph{58}) and the ClimbMix dataset (available at
\url{https://huggingface.co/datasets/karpathy/climbmix-400b-shuffle})
(\emph{59}). We constructed three disjoint subsets of 24M words from
each.

For comparability, all data were tokenized using a standard BPE
tokenizer trained on the full CHILDES dataset (all 24M). Surprisal was
computed as the mean per-word negative log likelihood over
\textasciitilde50 held-out validation contexts (last 128 tokens each)
for the 609 CDI words that were used in (\emph{34}) and that appeared in
the training data with sufficient frequency in the CHILDES held-out
contexts. Surprisal values for sigmoid fits were read out from the epoch
with the lowest validation loss.

For the computation of surprisal across training runs (``training''), we
trained models using 3 random seeds on the full dataset. For the
computation across different dataset sizes (``development''), the data
were segmented into 18 nested datasets ranging in size from .5M to the
full 24M. We then chose 10 random seed values and trained 10 models on
each dataset for a total of 180 models. For the computation of
variability across model runs (``composition''), we ran 8 random seed
values across 12 nested dataset sizes for three disjoint subsets of both
BabyLM and ClimbMix, resulting in 8 (seeds) x 12 (sizes) x 3 (subsets) x
2 (datasets) = 576 distinct models.

Models were trained for a total of 20 epochs with sequence length 1024
and batch size 8, using AdamW for optimization and learning rate of 1e-4
with a linear decay, no warmup, weight decay of 0, and no in-epoch
shuffling. Training was performed on NVIDIA A40 GPUs.

\subsection{Proportional Return on
Input}\label{proportional-return-on-input}

Proportional return on input was computed for models based on mean
surprisal \(L(D)\) of the CDI words after training on \(D\) word tokens
of input. This follows the form from (\emph{32}) of
\(L(D) = E + B\,D^{-\beta}\), with \(E\) the irreducible entropy floor
and \(B, \beta\) fit across budgets (here \(E \approx 2.94\) nats,
\(B \approx 312\), \(\beta \approx 0.32\), \(R^2 = 0.998\)). The excess
loss \(L(D) - E = B\,D^{-\beta}\) is the reducible part that training
removes; it falls as a power law of slope \(-\beta\). Marginal return is
the loss removed per \(e\)-fold of data,

\[R(D) = -\frac{\mathrm{d}L}{\mathrm{d}\ln D} = \beta\,\big(L(D) - E\big),\]

\noindent obtained by differentiating the scaling law. We compute the
ratio of this quantity to the excess loss, obtaining the proportional
return: the share of the remaining reducible loss removed per
\(e\)-fold, \(\gamma_{\text{LM}}(D) = \frac{R(D)}{L(D) - E} = \beta.\)
For a power law, this is constant:
\(\gamma_{\text{LM}} = \beta \approx 0.32\) at every budget,
recapitulating the scaling law.

We can compute the same quantity for children. We define a vocabulary
production loss, \(L_i(t) = \operatorname{mean}_j[-\log p_{ij}(t)]\),
whose floor is zero (a known word is produced, \(p \to 1\)). Its
marginal return is
\(R_i(t) = -\mathrm{d}L_i/\mathrm{d}\log t = \kappa_i\,\operatorname{mean}_j(1 - p_{ij})\),
so the proportional return is \[\gamma_i(t) = \frac{R_i(t)}{L_i(t)}
 = \kappa_i\,\frac{\operatorname{mean}_j(1 - p_{ij})}{\operatorname{mean}_j(-\log p_{ij})},\]
which rises toward the child's own \(\kappa_i\) as the vocabulary
saturates.

\subsection{Computing Acceleration in
LMs}\label{computing-acceleration-in-lms}

To compare acceleration in LMs and children, we link the \(\kappa\)
parameter from our accelerating accumulator model to the word-by-word
learning performance of LMs. We first focus on the slope for individual
words rather than individual children. Starting with the M3,
differentiating \(\eta_{ij}(t)\) (the probability of producing a
particular word) with respect to \(\log t\) gives \(\kappa_i\): the
logit-probability that a child produces a word rises linearly in log-age
with slope equal to the child's scaling exponent. Thus, the per-child
sigmoid slope \emph{is} \(\kappa_i\), by construction.

Next, to estimate the slope of acquisition of individual words,
(\emph{34}) compute surprisal for each word across LM training
checkpoints and then fit a four-parameter logistic,
\(s_w(x) = \ell_w + (u_w-\ell_w)/(1+\exp((x-m_w)/\mathrm{scale}_w))\),
in \(x = \log_{10}\) training data. Let
\(p_w = (s_w-\ell_w)/(u_w-\ell_w)\) be the fraction of that word's
surprisal reduction still outstanding. The LM slope is therefore
\(d\,\mathrm{logit}(p_w)/d\ln D = 1/(\mathrm{scale}_w \ln 10)\) --- a
logit per unit of natural-log input. That is the same as \(\kappa_i\)
for children, where the logit probability of producing a word rises in
log-age with slope \(\kappa_i\) and the outstanding fraction is
\(1-P(\text{produce})\).

\section{Supplementary Text}\label{supplementary-text}

\subsection{Comparison with Prior
Models}\label{comparison-with-prior-models}

Our accelerating accumulator model has not been described in the
literature in the exact form we use, but it is closely related to other
models. Here we describe these models as special cases of our model. We
write the latent ability of child \(i\) on word \(j\) at age \(t\) as

\[\eta_{i,j}(t) = \log r_i + \log \alpha_i + \kappa_i \log t - \delta_j,\]
where \(r_i\) and \(\alpha_i\) decompose the term \(\xi_i\) in our
model: \(r_i\) is the input rate for the child and \(\alpha_i\) is their
learning efficiency. Several prior accumulator models can then be seen
as special cases in which one or more of these components is fixed:

\begin{itemize}
\tightlist
\item
  (\emph{20}) is a pure accumulator in which \(\kappa=1\) and children
  do not differ from one another
  (\(\sigma_{\alpha}=\sigma_{\kappa}=0\)), so both input rate and
  efficiency reduce to a constant, corresponding to our M0.
\item
  (\emph{8}) introduces a model in which ability grows as \(t^{D+1}\),
  creating acceleration. This exponent is exactly our \(\kappa\), and
  this model is equivalent to our M1.
\item
  (\emph{9}) present our M2: they allow \(\alpha\) to vary across
  individuals, but not \(\kappa\).
\end{itemize}

In addition, several other models in the literature can be described as
variants of this framework:

\begin{itemize}
\tightlist
\item
  (\emph{60}) augment the pure M0 accumulator with a ``leverage'' term,
  in which words learned earlier facilitate the learning of later words.
  Their central result is a negative one: leverage changes the shape and
  timing of vocabulary growth, but it cannot create acceleration that is
  not already present in the distribution of word difficulties.
\item
  (\emph{21}) define a pure accumulator model with \(\kappa=1\), but
  with an additional parameter relating to the speed of per-word
  accumulation. This model maps onto a combination of our M0 and the
  2-parameter logistic variant that we fit below.
\end{itemize}

\subsection{Non-identifiability in Cross-Sectional
Data}\label{non-identifiability-in-cross-sectional-data}

\begin{figure}[H]

\centering{

\includegraphics[width=1\linewidth,height=\textheight,keepaspectratio]{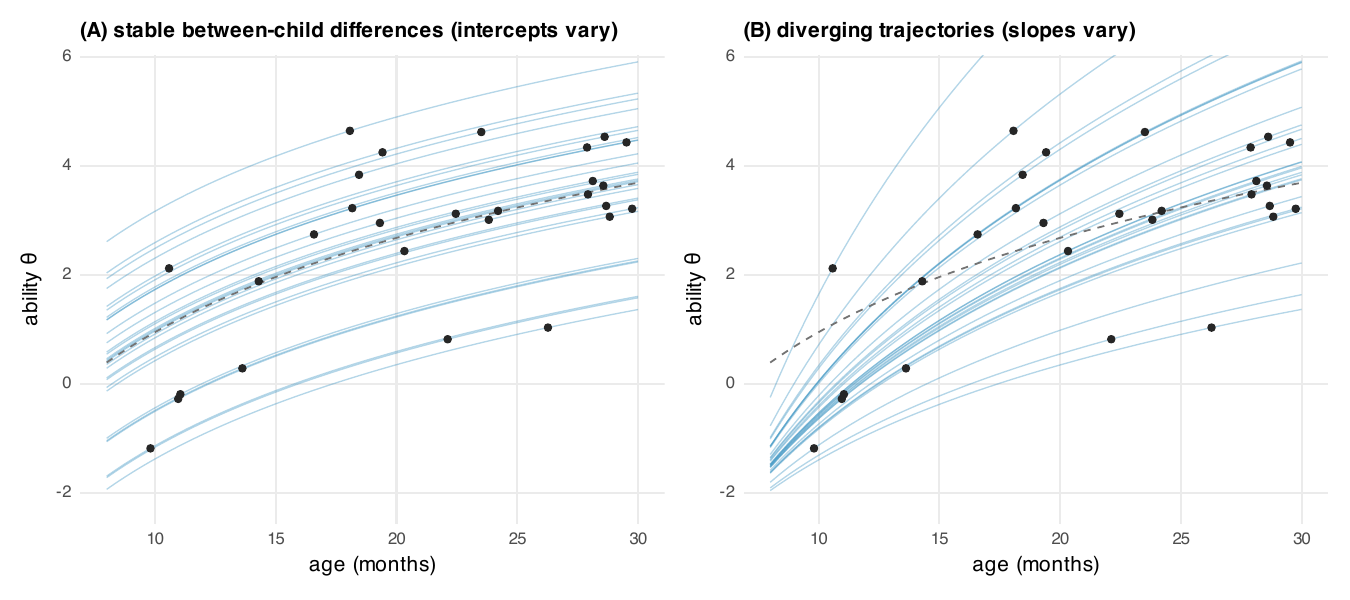}

}

\caption{\label{fig-cross-sectional-confound}Non-identifiability of
growth models in cross-sectional data. Panels show the same simulated
data, reflecting the possibility of stable between-child intercept
differences (A) vs.~varying growth slopes (B).}

\end{figure}

Since acceleration is a property of change over time, a single
cross-sectional observation from an individual child cannot identify the
child's acceleration. As Figure~\ref{fig-cross-sectional-confound}
shows, a single observation from a child could be accounted for by
variation in either \(\xi_i\) (efficiency) or \(\kappa_i\)
(acceleration).

\subsection{Datasets}\label{datasets}

Table \ref{tbl-datasets} gives the characteristics of each dataset used.
In the main text we focus on children with three or more administrations
(for more precise estimates of \(\kappa\)), but below we also confirm
the robustness of our claims including children with two or more
administrations.

\begin{table}

\caption{\label{tbl-datasets}Characteristics of all datasets used in the
paper. N and ages reflect the analysis samples (post-exclusion) for the
two exclusion criteria (3+ and 2+ administrations). Adm/Chi gives the
median administrations per child.}

\centering{

\centering\begingroup\fontsize{9}{11}\selectfont

\begin{tabular}{llrrrrrr}
\toprule
\multicolumn{4}{c}{ } & \multicolumn{3}{c}{Age (months)} & \multicolumn{1}{c}{ } \\
\cmidrule(l{3pt}r{3pt}){5-7}
Citation & Language & N (3+/2+) & Admins & Mean & Min & Max & Adm/Chi\\
\midrule
Thal (Wordbank) & English (American) & 639 /   653 & 1,919 & 19.0 & 12 & 29 & 3\\
Smith (Wordbank) & English (American) & 152 /   316 & 607 & 22.2 & 16 & 30 & 4\\
Marchman (Wordbank) & English (American) & 162 / 2,136 & 499 & 20.3 & 8 & 30 & 3\\
Simonsen et al. (2014) & Norwegian & 792 / 1,630 & 4,770 & 24.1 & 8 & 36 & 5\\
Hagihara et al. (2023) & Japanese & 96 /   187 & 289 & 16.5 & 11 & 33 & 3\\
\bottomrule
\end{tabular}
\endgroup{}

}

\end{table}%

\begin{figure}[H]

\centering{

\includegraphics[width=1\linewidth,height=\textheight,keepaspectratio]{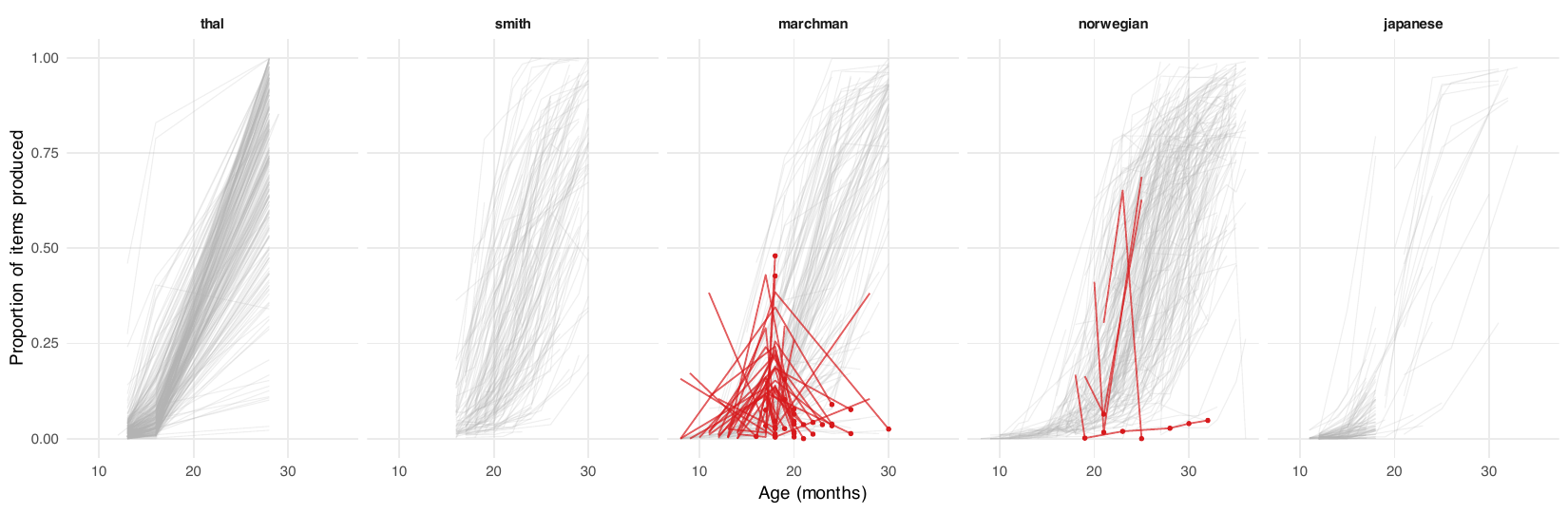}

}

\caption{\label{fig-qc}Data-quality exclusions. Red lines show children
excluded from the analysis entirely, with red dots marking flagged
administrations; gray lines are a random sample of retained children,
shown for reference.}

\end{figure}

Fig. \ref{fig-qc} shows the full spaghetti plots for each dataset,
highlighting in red those observations that failed our quality check and
were excluded. We speculate that these observations are due to either
data entry issues or faulty correspondences between observations. Such
issues occur frequently in archival datasets and affect only a small
portion of our data.

\subsection{Full Model Comparison
Results}\label{full-model-comparison-results}

Table \ref{tbl-loo} gives the full leave-one-out model comparison
results across models for both the dataset with 3+ administrations/child
and the larger dataset with 2+ administrations/child.

\begin{table}

\caption{\label{tbl-loo}Bayesian leave-one-out (LOO) model comparison.
Each cell is the difference in expected log predictive density (ELPD)
relative to the best model (M3), with the standard error of the
difference in parentheses; more negative is worse.}

\centering{

\centering\begingroup\fontsize{8}{10}\selectfont

\resizebox{\ifdim\width>\linewidth\linewidth\else\width\fi}{!}{
\begin{tabular}{lrrrrr}
\toprule
\multicolumn{1}{c}{ } & \multicolumn{5}{c}{$\Delta$ELPD vs M3 (SE)} \\
\cmidrule(l{3pt}r{3pt}){2-6}
Model & Thal & Smith & Marchman & Norwegian & Japanese\\
\midrule
\addlinespace[0.3em]
\multicolumn{6}{l}{\textbf{$\geq$3 administrations (main text)}}\\
\hspace{1em}M0. accumulator ($\kappa=1$) & -154,678 (350) & -102,675 (340) & -83,954 (285) & -130,396 (369) & -24,628 (186)\\
\hspace{1em}M1. + acceleration & -46,214 (289) & -62,282 (309) & -40,868 (258) & -70,888 (336) & -7,923 (126)\\
\hspace{1em}M2. + efficiency var. & -7,880 (133) & -6,542 (121) & -4,878 (110) & -5,702 (111) & -761 ( 40)\\
\hspace{1em}M3. + acceleration var. & 0 & 0 & 0 & 0 & \vphantom{1} 0\\
\addlinespace[0.3em]
\multicolumn{6}{l}{\textbf{$\geq$2 administrations (full longitudinal)}}\\
\hspace{1em}M0. accumulator ($\kappa=1$) & -155,002 (351) & -120,701 (372) & -125,833 (355) & -131,793 (373) & -53,239 (260)\\
\hspace{1em}M1. + acceleration & -45,613 (287) & -75,934 (343) & -66,770 (332) & -75,942 (346) & -21,922 (197)\\
\hspace{1em}M2. + efficiency var. & -7,723 (132) & -7,795 (129) & -14,004 (195) & -6,047 (115) & -1,919 ( 65)\\
\hspace{1em}M3. + acceleration var. & 0 & 0 & 0 & 0 & 0\\
\bottomrule
\end{tabular}}
\endgroup{}

}

\end{table}%

\subsection{Individual-level
Cross-validation}\label{individual-level-cross-validation}

The model comparisons above report ELPD LOO values for held-out data
from individual item-responses, which are nested within children. To
assess the value of M3 (the accelerating accumulator) in making fully
out-of-sample predictions, we designed a prospective test. As our data,
we selected a subset of the Norwegian sample that included 371 children
with \(\geq 6\) longitudinal CDIs. We then fit models to the child's
most recent \(k\) observations and used these to predict the child's
next administration (3 months later). This design avoids confounding
data density with the gap between the fitted and predicted observations.

Results from this analysis are shown in Table~\ref{tbl-cv-depth}. The
result is very sensitive to the value of \(k\) used. With \(k=2\), the
fixed slope model (M2) clearly predicts better because the per-child
slope model (M3) fully parameterizes the two observations per child and
absorbs measurement error as a result. This effect decreases as \(k\)
increases until the two models are tied by \(k=5\). We expect that with
even denser data, a separate slope parameter would lead to individually
better fits, but in practice the single slope model makes acceptable
predictions.

Importantly, because both M2 and M3 contain a population acceleration
parameter, this analysis does not test for the presence of acceleration.
Instead it suggests that measuring the precise amount of acceleration
each child is undergoing requires sustained, precise longitudinal
measurement.

\begin{table}

\caption{\label{tbl-cv-depth}Prospective cross-validation for Norwegian
children with at least six administrations (n = 371). Each row describes
a model fit to k administrations and scored on the child's held-out
final sitting. dELPD/child is the per-child difference in expected log
predictive density, full model minus shared-slope model; negative favors
the shared slope. sigma\_b is the between-child acceleration SD in the
training fit, against a full-data value of 5.65.}

\centering{

\centering\begingroup\fontsize{9}{11}\selectfont

\begin{tabular}{rrrrrrr}
\toprule
$k$ & horizon (mo) & $\Delta$ELPD/child & SE & $z$ & \% children better & $\sigma_b$\\
\midrule
2 & 3 & -32.41 & 6.77 & -4.8 & 39 & 12.56\\
3 & 3 & -5.32 & 4.10 & -1.3 & 49 & 7.86\\
4 & 3 & -0.33 & 3.64 & -0.1 & 48 & 6.20\\
5 & 3 & +0.02 & 3.51 & +0.0 & 51 & 5.17\\
\bottomrule
\end{tabular}
\endgroup{}

}

\end{table}%

We next tested whether acceleration plays an identifiable role in
predicting individual children's trajectories at all. We therefore fit
M2\(_0\), a variant of M2 with the acceleration exponent constrained to
the pure accumulator value (\(\kappa = 1\)) --- rather than the
population acceleration rate -- while retaining the per-child efficiency
term \(\xi_i\). The comparison of M2 and M2\(_0\) thus tests the
hypothesis that acceleration is predictive of individual trajectories.
Table \ref{tbl-cv-headline} shows the results of this comparison. M2,
the accelerating model, predicted held-out administrations better at
every depth.

\begin{table}

\caption{\label{tbl-cv-headline}Prospective cross-validation of
acceleration, on the same Norwegian children and held-out
administrations as Table~\ref{tbl-cv-depth}. M2 retains a free
population acceleration exponent; M2-zero removes it. dELPD/child is the
per-child difference in expected log predictive density, M2 minus
M2-zero; positive favors acceleration.}

\centering{

\centering\begingroup\fontsize{9}{11}\selectfont

\begin{tabular}{rrrrrr}
\toprule
$k$ & horizon (mo) & $\Delta$ELPD/child & SE & $z$ & \% children better\\
\midrule
2 & 3 & +91.87 & 6.97 & +13.2 & 84\\
3 & 3 & +143.21 & 8.90 & +16.1 & 88\\
4 & 3 & +197.27 & 10.33 & +19.1 & 90\\
5 & 3 & +248.10 & 11.22 & +22.1 & 92\\
\bottomrule
\end{tabular}
\endgroup{}

}

\end{table}%

\subsection{Comparison to a Non-Parametric Quantile
Model}\label{comparison-to-a-non-parametric-quantile-model}

To test whether the accelerating accumulator appropriately fits the
distribution of growth curves across children, we compared against a
strong baseline: the flexible non-parametric GAMLSS family that is used
to produce normative curves for the CDI (\emph{19}). Specifically, for
each dataset we fit a GAMLSS beta-regression of the proportion of words
produced by the child's age. We used a penalized spline function on the
mean and on the scales.

\begin{figure}[H]

\centering{

\includegraphics[width=1\linewidth,height=\textheight,keepaspectratio]{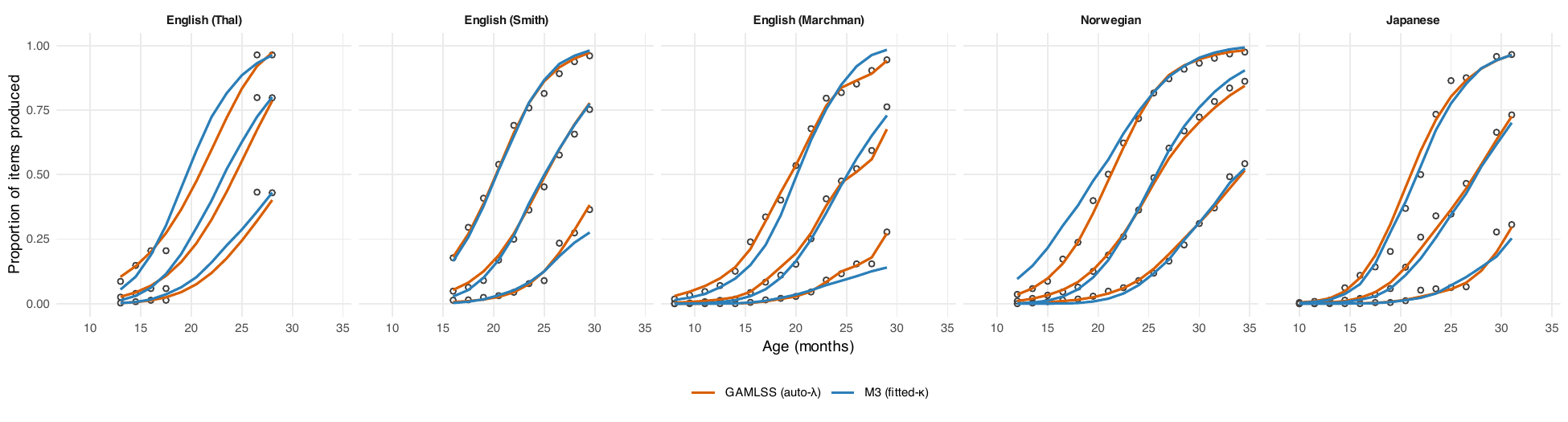}

}

\caption{\label{fig-gamlss-overlay}M3 fits (blue, drawn from fitted
per-child acceleration) vs.~non-parametric GAMLSS beta-regression
(orange), with empirical 10th/50th/90th percentiles (open circles).
Lines are the 10/50/90 percentiles of each model. Color denotes model.}

\end{figure}

Fig. \ref{fig-gamlss-overlay} shows that across all five datasets the M3
fit (blue) tracks both the non-parametric GAMLSS fit (orange) and the
empirical quantiles (circles). Thus, the accelerating accumulator is a
faithful, low-dimensional compression of the data's quantile structure.

\subsection{Robustness Across Model and Data
Settings}\label{robustness-across-model-and-data-settings}

In the main text, we report fits to children with 3+ administrations
with separate models fit to each dataset, but our parameter estimates
were substantially similar across two other settings: first, separate
fits to children with 2+ administrations, and second, a full
hierarchical model fit to children with 3+ administrations. Table
\ref{tbl-settings} shows LOO fits and the robustness of the \(\kappa\)
estimates across these different settings.

The pooled model was structured similarly to the by-dataset models, but
children and words were modeled as nested within their own dataset. Each
dataset had its own mean \(\xi\) and \(\kappa\) values. Dataset-level
standard deviations on efficiency and acceleration (\(\sigma_a\) and
\(\sigma_b\)) were distributed as
\(\log \sigma_{a[d]} \sim \mathcal N(m_a, s_a)\) and
\(\log \sigma_{b[d]} \sim \mathcal N(m_b, s_b)\), so a dataset with few
children had its estimated spread drawn toward the typical value across
datasets, while a dataset with many children was left essentially
unconstrained. The item-difficulty distribution was estimated separately
within each dataset, since the different datasets reflect data from
different languages. Overall, the pooled model reproduced the parameters
of the independent fits closely.

\begin{table}

\caption{\label{tbl-settings}For each dataset and setting: number of
children (N), population acceleration (\(\kappa\)), between-child
acceleration SD (\(\sigma_b\)), and the per-observation LOO advantage of
the full model M3 over M2 (\(\Delta\)ELPD per item response).}

\centering{

\centering\begingroup\fontsize{9}{11}\selectfont

\begin{tabular}{lrrrr}
\toprule
Setting & N & $\kappa$ & $\sigma_b$ & $\Delta$ELPD/obs (M2$\to$M3)\\
\midrule
\addlinespace[0.3em]
\multicolumn{5}{l}{\textbf{English (Thal)}}\\
\hspace{1em}$\geq$2 separate & 653 & 11.5 & 3.19 & 0.0154\\
\hspace{1em}$\geq$3 separate (main text) & 639 & 11.5 & 3.19 & 0.0158\\
\hspace{1em}pooled ($\geq$3) & 639 & 11.5 & 3.22 [3.07, 3.38] & ---\\
\addlinespace[0.3em]
\multicolumn{5}{l}{\textbf{English (Smith)}}\\
\hspace{1em}$\geq$2 separate & 316 & 12.8 & 8.06 & 0.0156\\
\hspace{1em}$\geq$3 separate (main text) & 152 & 12.9 & 5.14 & 0.0159\\
\hspace{1em}pooled ($\geq$3) & 152 & 12.8 & 5.13 [4.67, 5.63] & ---\\
\addlinespace[0.3em]
\multicolumn{5}{l}{\textbf{English (Marchman)}}\\
\hspace{1em}$\geq$2 separate & 2,136 & 10.6 & 6.89 & 0.0280\\
\hspace{1em}$\geq$3 separate (main text) & 162 & 10.6 & 6.60 & 0.0164\\
\hspace{1em}pooled ($\geq$3) & 162 & 10.4 & 6.51 [5.87, 7.30] & ---\\
\addlinespace[0.3em]
\multicolumn{5}{l}{\textbf{Norwegian}}\\
\hspace{1em}$\geq$2 separate & 1,630 & 12.9 & 7.63 & 0.0121\\
\hspace{1em}$\geq$3 separate (main text) & 792 & 13.3 & 5.65 & 0.0114\\
\hspace{1em}pooled ($\geq$3) & 792 & 13.2 & 5.65 [5.40, 5.93] & ---\\
\addlinespace[0.3em]
\multicolumn{5}{l}{\textbf{Japanese}}\\
\hspace{1em}$\geq$2 separate & 187 & 12.0 & 5.46 & 0.0074\\
\hspace{1em}$\geq$3 separate (main text) & 96 & 11.6 & 5.45 & 0.0059\\
\hspace{1em}pooled ($\geq$3) & 96 & 11.5 & 5.40 [4.74, 6.25] & ---\\
\bottomrule
\end{tabular}
\endgroup{}

}

\end{table}%

\subsection{Sensitivity to Data-Quality
Exclusions}\label{sensitivity-to-data-quality-exclusions}

To ensure that the exclusion filter we used in the main text does not
result in significant changes to our estimates, we refit the model at
four settings: with the filter disabled entirely, at a loose threshold,
at the reported threshold, and at a tight one
(Table~\ref{tbl-qc-sensitivity}). These refits apply only to two
datasets, English (Marchman) and Norwegian. The exclusion filter does
not result in major changes to \(\kappa\) estimates in either case.

\begin{table}

\caption{\label{tbl-qc-sensitivity}Population acceleration and
between-child acceleration SD by data-quality filter setting, for the
two datasets in which the filter removes observations. The loose filter
removes children with a decline of more than 40\% below a child's
running peak or a rise of more than 60 percentage points per month; the
main filter uses a decline of 25\% and a rise of 40\%, as reported in
the Methods; and the tight filter uses 15\% and 25\%.}

\centering{

\centering\begingroup\fontsize{9}{11}\selectfont

\begin{tabular}{lrrr}
\toprule
Filter & Removed & $\kappa$ & $\sigma_b$\\
\midrule
\addlinespace[0.3em]
\multicolumn{4}{l}{\textbf{English (Marchman)}}\\
\hspace{1em}none & 0 & 9.37 & 8.33\\
\hspace{1em}loose & 29 & 11.17 & 8.30\\
\hspace{1em}main & 32 & 10.65 & 6.60\\
\hspace{1em}tight & 34 & 10.44 & 5.55\\
\addlinespace[0.3em]
\multicolumn{4}{l}{\textbf{Norwegian}}\\
\hspace{1em}none & 0 & 13.05 & 5.77\\
\hspace{1em}loose & 36 & 13.19 & 5.64\\
\hspace{1em}main & 46 & 13.25 & 5.65\\
\hspace{1em}tight & 55 & 13.22 & 5.34\\
\bottomrule
\end{tabular}
\endgroup{}

}

\end{table}%

\subsection{Linear vs.~Logarithmic
Age}\label{linear-vs.-logarithmic-age}

We additionally fit M3 variants with linear, rather than logarithmic,
growth in acceleration over time. Table \ref{tbl-loglin} shows that the
linear model was dispreferred across all five datasets.

\begin{table}

\caption{\label{tbl-loglin}Log-age vs linear-age accumulation. LOO ELPD
advantage of the log-age model (M3) over its linear-age counterpart,
with SE; positive favors log-age.}

\centering{

\centering\begingroup\fontsize{9}{11}\selectfont

\begin{tabular}{lr}
\toprule
Dataset & $\Delta$ELPD (log $-$ linear)\\
\midrule
Thal & 267 (31)\\
Smith & 249 (23)\\
Marchman & 281 (26)\\
Norwegian & 323 (39)\\
Japanese & 39 ( 8)\\
\bottomrule
\end{tabular}
\endgroup{}

}

\end{table}%

\subsection{2PL Model Comparison}\label{pl-model-comparison}

In the main analysis we use a Rasch (1PL) item model, in which every
word has a difficulty but all words have equal discrimination between
children. (\emph{61}) found that a 2PL model, which adds a per-word
discrimination \(\lambda_j\), was preferred over the 1PL for CDI data.
We therefore refit M3 with a per-word discrimination parameter.

Table \ref{tbl-2pl} shows the results of this comparison. The 2PL does
genuinely fit better, with substantial differences in LOO ELPD for every
language. However, model results are otherwise unchanged. Estimates of
\(\kappa\) increase somewhat, but they are not directly comparable
across item models. In normal IRT models, ability scores are constrained
to a standard normal distribution, but in our model, abilities grow over
time so there is no single standard scale, and the addition of the
discrimination parameter can change the overall scale. For this reason,
we report \(\log \kappa\), since a pure change of units would be a
constant shift. In addition, across individual words, we find that the
predicted age of acquisition (the age at which the word is predicted to
have a 50\% probability of being known) is closely aligned, suggesting
that the models are highly comparable.

Critically, \(\delta\) (difficulty) and \(\lambda\) (discrimination)
correlations were small to modest. If harder words were substantially
more discriminating, this correlation could inflate estimates of
\(\kappa\) in the 1PL model; in practice we find little support for this
hypothesis.

\begin{table}

\caption{\label{tbl-2pl}Rasch (1PL) versus two-parameter logistic (2PL)
item models, M3 refit on the same data. Discrimination: SD of
\(\log \lambda_j\), its 10th-90th percentile range on the natural scale,
and its correlation with item difficulty (90\% interval). Acceleration:
population \(\kappa\) under each item model and the difference in
\(\log \kappa\). Age of acquisition: correlation and median absolute
difference in per-word \(t_{50}\), which is invariant to the
discrimination scale. LOO: expected log predictive density of the 1PL
relative to the 2PL, with SE; negative favors the 2PL.}

\centering{

\centering\begingroup\fontsize{7}{9}\selectfont

\resizebox{\ifdim\width>\linewidth\linewidth\else\width\fi}{!}{
\begin{tabular}{lrrrrrrrrrrr}
\toprule
\multicolumn{1}{c}{ } & \multicolumn{3}{c}{Discrimination} & \multicolumn{3}{c}{Acceleration $\kappa$} & \multicolumn{2}{c}{Word slopes $\lambda_j\kappa_i$} & \multicolumn{2}{c}{Age of acquisition} & \multicolumn{1}{c}{LOO} \\
\cmidrule(l{3pt}r{3pt}){2-4} \cmidrule(l{3pt}r{3pt}){5-7} \cmidrule(l{3pt}r{3pt}){8-9} \cmidrule(l{3pt}r{3pt}){10-11} \cmidrule(l{3pt}r{3pt}){12-12}
Dataset & SD $\log\lambda$ & $\lambda$ p10--p90 & $r(\lambda,\delta)$ & 1PL & 2PL & $\Delta\log\kappa$ & med. & \% $>1$ & $r(t_{50})$ & med.\ $|\Delta t_{50}|$ & $\Delta$ELPD (SE)\\
\midrule
English (Thal) & 0.25 & 0.73--1.34 & +0.13 [+0.11, +0.15] & 11.5 & 12.4 & +0.072 & 12.2 & 99.8 & 0.999 & 0.13 & -2,897 ( 82)\\
English (Smith) & 0.27 & 0.71--1.37 & +0.10 [+0.07, +0.13] & 12.9 & 13.7 & +0.053 & 12.9 & 99.3 & 0.996 & 0.13 & -2,726 ( 79)\\
English (Marchman) & 0.24 & 0.74--1.33 & +0.05 [+0.02, +0.09] & 10.6 & 11.6 & +0.089 & 11.1 & 98.4 & 0.998 & 0.25 & -1,412 ( 59)\\
Norwegian & 0.29 & 0.70--1.36 & +0.22 [+0.21, +0.23] & 13.3 & 14.1 & +0.065 & 13.7 & 99.4 & 0.995 & 0.16 & -4,426 (108)\\
Japanese & 0.34 & 0.65--1.51 & +0.07 [+0.02, +0.12] & 11.6 & 15.3 & +0.279 & 14.7 & 100.0 & 0.986 & 1.47 & -641 ( 41)\\
\bottomrule
\end{tabular}}
\endgroup{}

}

\end{table}%

\subsection{Sensitivity to Item
Selection}\label{sensitivity-to-item-selection}

\begin{figure}[H]

\centering{

\includegraphics[width=1\linewidth,height=\textheight,keepaspectratio]{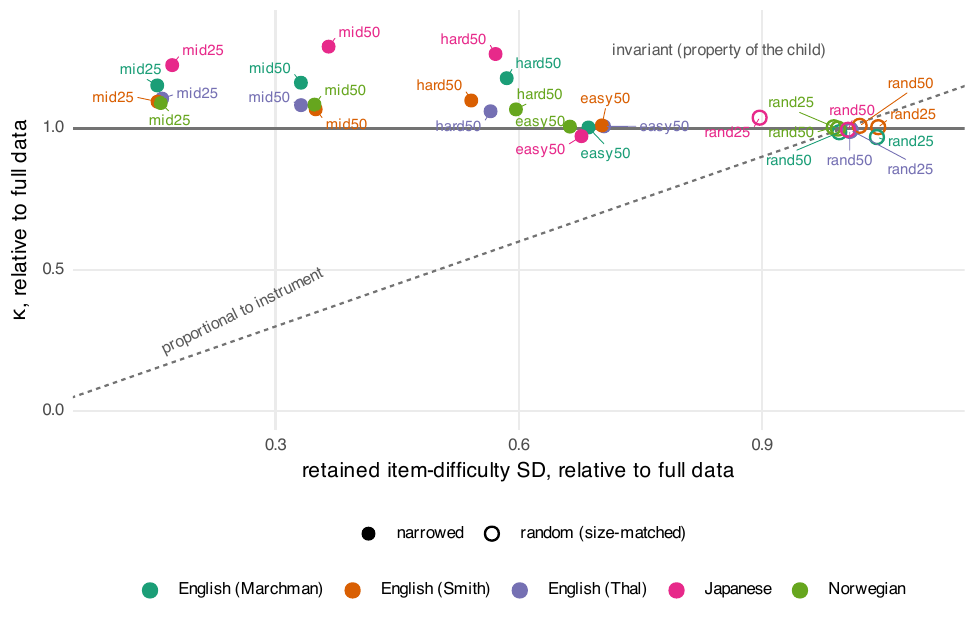}

}

\caption{\label{fig-item-subset}Acceleration under deliberately narrowed
item sets, plotted as the ratio to the full-data fit against the ratio
of retained item-difficulty spread. Each point is one refit on the same
children and administrations. If acceleration rescaled with the
instrument, points would fall on the diagonal; if it is a property of
the children, they fall on the horizontal line at one. Open points are
random subsets of matched size, which discard the same number of items
without narrowing the spread.}

\end{figure}

One potential concern about our \(\kappa\) estimates is that they could
be influenced by the composition of the CDI instrument. Could a broader
spread of item difficulties result in a larger \(\kappa\) estimate? To
address this issue, we refit M3 on the same children and same
administrations, but with the item set narrowed. We fit models under
three sets of conditions: refitting the middle 50\% and middle 25\% of
items (to narrow the width of the distribution), refitting the easiest
and hardest halves separately (to shift the difficulty distribution),
and subsetting randomly to 50\% and 25\% of items. If \(\kappa\) scales
as a function of the difficulty distribution (\(\sigma_\delta\)),
recovered \(\kappa\) values should scale with the standard deviation of
the retained items; if not, \(\kappa\) should be relatively invariant.
Fig. \ref{fig-item-subset} shows the results of this analysis. Overall,
\(\kappa\) values were invariant to retained item variability.

\subsection{Variation in Acceleration}\label{variation-in-acceleration}

\begin{figure}[H]

\centering{

\includegraphics[width=1\linewidth,height=\textheight,keepaspectratio]{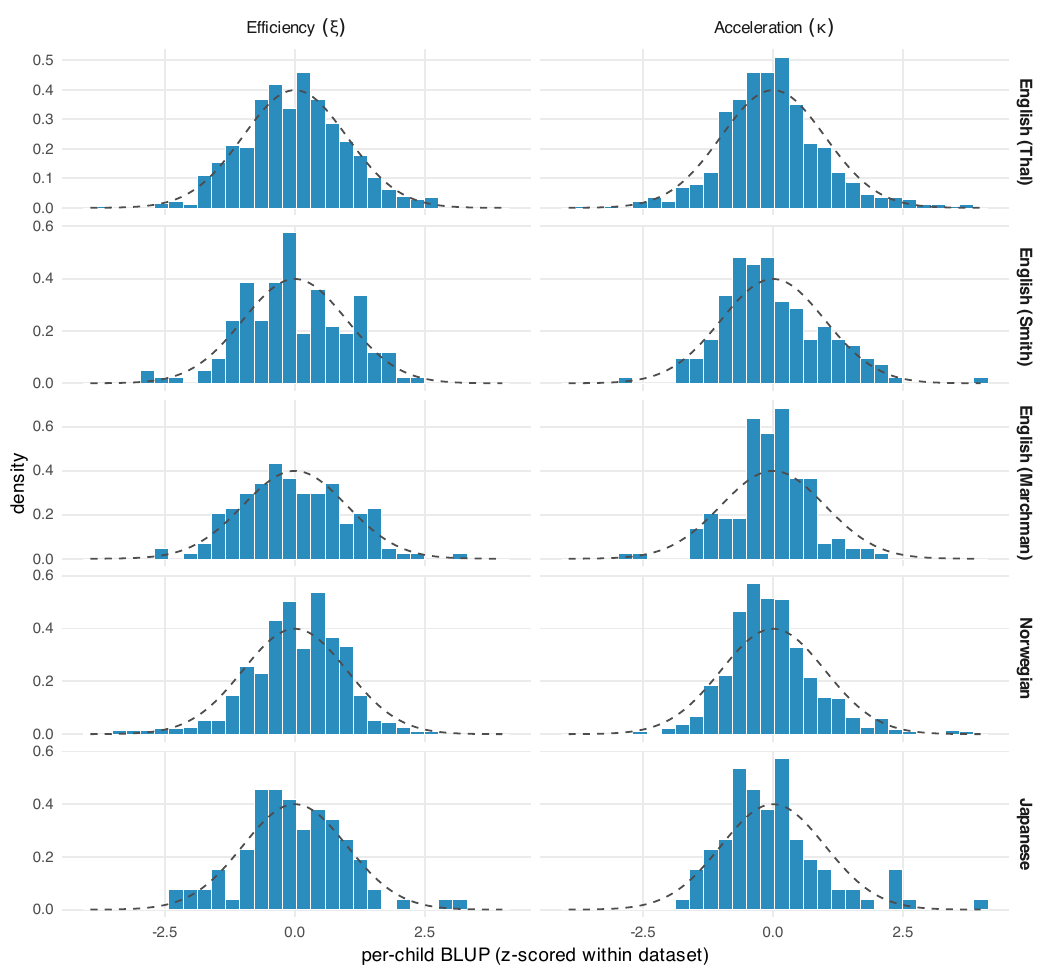}

}

\caption{\label{fig-blup-hist}Histograms of per-child BLUP estimates,
z-scored within each dataset, for both efficiency and acceleration
parameters. Dashed lines indicate the standard normal reference
distribution.}

\end{figure}

To explore the distribution of individual children's fitted parameters,
we show the best linear unbiased predictor (BLUP) for each child
(Figure~\ref{fig-blup-hist}) as well as the random effects of the fitted
models (Table~\ref{tbl-ranef}). While the distribution appears generally
normal, we do not give a strong interpretation as this shape may in part
be due to the normal form of the model's random effect structure.
However, across all datasets, acceleration appeared right-skewed.

\begin{table}

\caption{\label{tbl-ranef}Random-effect parameters for M3 fits, by
dataset: population acceleration \(\kappa\), between-child SD in
efficiency (\(\sigma_a\)) and acceleration (\(\sigma_b\)), their
correlation (\(\rho\)), and the percentage of children estimated above
the pure-accumulator value \(\kappa = 1\). Brackets give 90\% posterior
intervals.}

\centering{

\centering\begingroup\fontsize{8}{10}\selectfont

\resizebox{\ifdim\width>\linewidth\linewidth\else\width\fi}{!}{
\begin{tabular}{lrrrrrr}
\toprule
Dataset & N & $\kappa$ & $\sigma_a$ & $\sigma_b$ & $\rho$ & \% $\kappa_i>1$\\
\midrule
English (Thal) & 639 & 11.5 [11.5, 11.6] & 1.56 [1.49, 1.64] & 3.19 [3.05, 3.35] & +0.10 [+0.03, +0.17] & 99.8\\
English (Smith) & 152 & 12.9 [12.8, 13.1] & 2.08 [1.89, 2.31] & 5.14 [4.69, 5.68] & -0.26 [-0.39, -0.13] & 99.3\\
English (Marchman) & 162 & 10.6 [10.5, 10.8] & 1.80 [1.65, 1.99] & 6.60 [5.90, 7.32] & +0.17 [+0.04, +0.29] & 97.5\\
Norwegian & 792 & 13.3 [13.2, 13.3] & 2.71 [2.60, 2.83] & 5.65 [5.41, 5.88] & -0.62 [-0.66, -0.58] & 99.5\\
Japanese & 96 & 11.6 [11.1, 12.1] & 1.95 [1.71, 2.25] & 5.45 [4.74, 6.32] & +0.26 [+0.07, +0.43] & 100.0\\
\bottomrule
\end{tabular}}
\endgroup{}

}

\end{table}%

\subsection{Input Estimation}\label{input-estimation}

We make use of estimates of children's language input in both Fig. 2 and
Fig. 3C. We calculate average language input per month by estimating
words per hour and multiply by an estimated number of waking hours per
month. Word tokens per hour estimates come from 42 children with
manually transcribed data in two prior studies (\emph{56}, \emph{57},
estimates used also in \emph{9}). Across these studies, the pooled
geometric mean is \(\exp(7.338) \approx 1{,}537\) word tokens per hour
of child-directed (any-adult) speech, with between-child standard
deviation of log rate \(\sigma_r = 0.534\). We multiply these estimates
by an assumed 12 waking exposure-hours per day to get a monthly average
number of hours (\(12 \times 30.4 \approx 365\)), giving
\(1{,}537 \times 365 \approx 561\text{K}\) tokens per month. A child's
average cumulative input at age \(t\) months is then
\(t \times 561\text{K}\) under a constant-rate assumption.

In Fig. 3C, we give uncertainty bands on representative children based
on \(\pm 1\) SD of the between-child log input rate. A child one
standard deviation above (below) the median rate reaches the same age
having heard \(\times 1.71\) (\(\div 1.71\)) as many tokens. This
results in a horizontal shift of the child's curve along the log-token
axis.

\begin{figure}[H]

\centering{

\includegraphics[width=1\linewidth,height=\textheight,keepaspectratio]{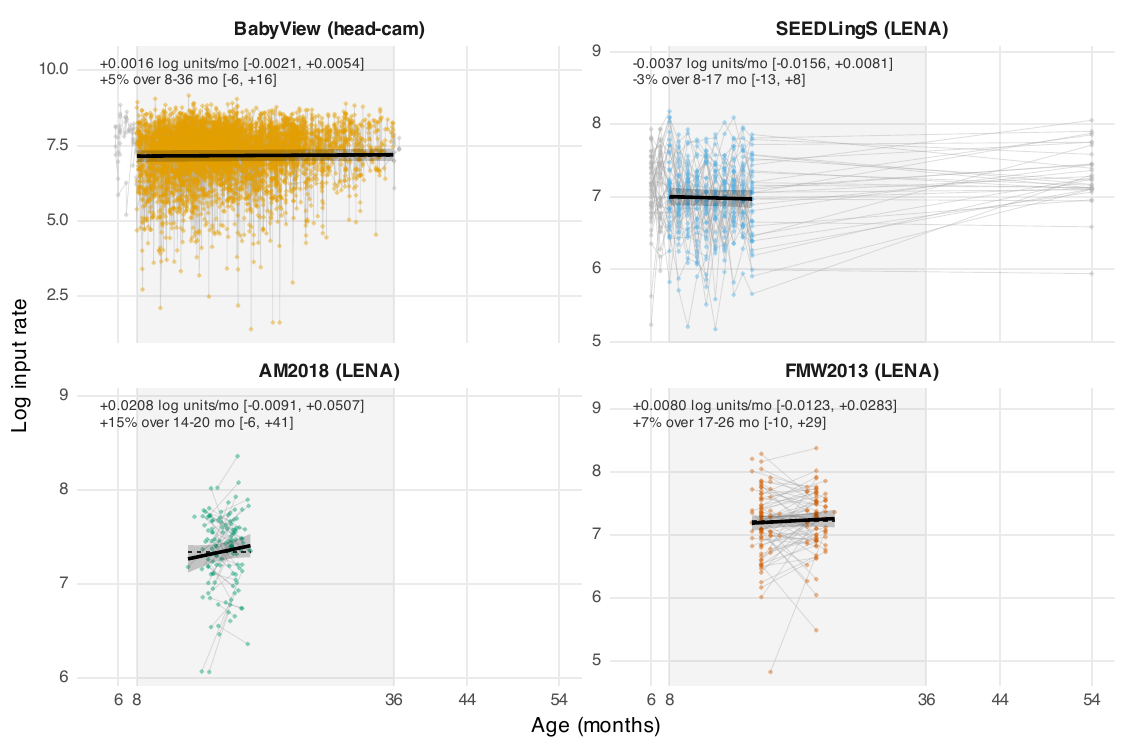}

}

\caption{\label{fig-input-age}Input rate does not change appreciably
across development. Each panel plots log adult input rate against child
age, with one point per recording, faint gray lines joining recordings
from the same child, and the fitted within-child trend (black) with its
95\% confidence band. Dashed lines show zero slope. Trends are fitted
over 8-36 months (shaded), the range over which the constant-rate
assumption is applied; gray points fall outside that window and are
excluded from the fit.}

\end{figure}

Our calibration computations assume that input is constant across age.
To provide empirical verification, we used data from four pre-existing
datasets: (\emph{62}), (\emph{63}) (BabyView), (\emph{64}), (\emph{65})
(SEEDLings). These studies measured children's input using either LENA
recorders or head-mounted cameras; we computed total words heard per
hour in each dataset for each recording. Fig. \ref{fig-input-age} shows
speech rate per hour in these datasets. Slopes over age for each were
shallow and non-significant.

\subsection{Age of Acquisition in the Pure Accumulator
Model}\label{age-of-acquisition-in-the-pure-accumulator-model}

Figure 2 in the main text converts each word's fitted difficulty into an
estimated number of cumulative exposures at the age it is acquired.
Since later-acquired words are also rarer, here we ask whether that
frequency change would be sufficient to produce the observed
relationship between cumulative exposure and age of acquisition.

\begin{figure}[H]

\centering{

\includegraphics[width=1\linewidth,height=\textheight,keepaspectratio]{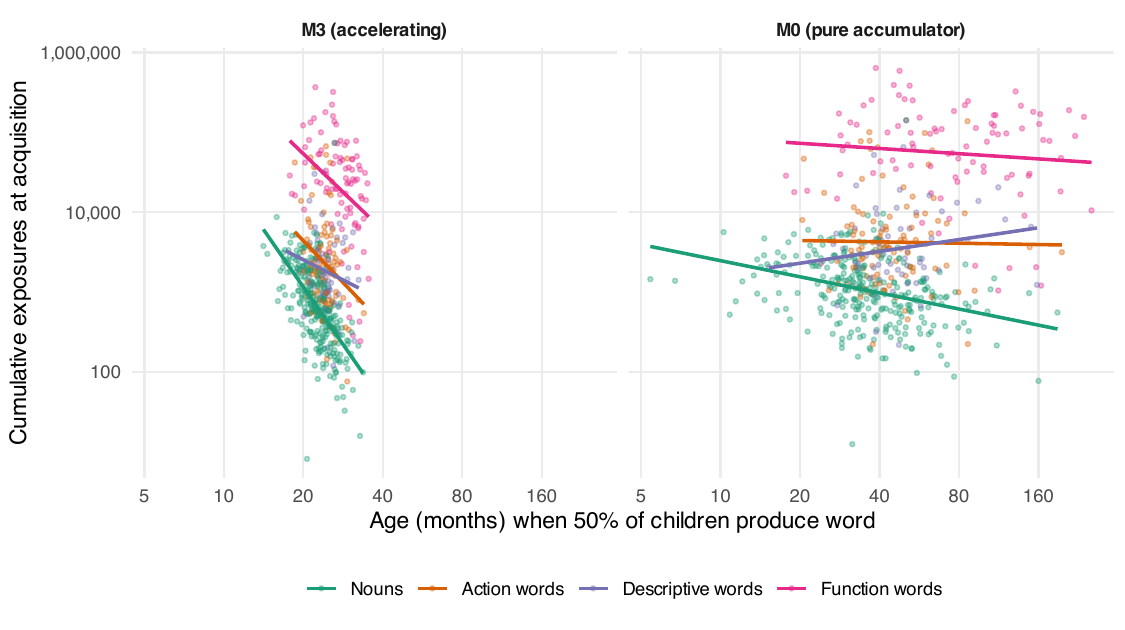}

}

\caption{\label{fig-aoa-m0}Estimated cumulative exposures at age of
acquisition under the accelerating accumulator (M3, left, reproduced
from main text) and the fitted pure accumulator (M0, right), using
identical corpus frequencies. Points are words, colored by class, with
per-class linear fits.}

\end{figure}

Fig. \ref{fig-aoa-m0} shows the same relationship, but comparing age of
acquisition estimates between M3 (the accelerating accumulator) and M0
(the pure accumulator). The pure accumulator does not reproduce either
the scale or the slope of the relationship (Table~\ref{tbl-aoa-m0}).
Within every word class, the slope of the relationship is substantially
shallower in M0 than M3, and the M0 forces the fitted difficulties to
spread acquisition across 5 to 254 months, against 14 to 35 months for
the accelerating model and 8 to 36 months of actual observation.

\begin{table}

\caption{\label{tbl-aoa-m0}Slope of estimated cumulative exposures with
respect to age of acquisition, by word class, under each item model.
Negative values mean fewer exposures are needed for later-acquired
words.}

\centering{

\centering\begingroup\fontsize{9}{11}\selectfont

\begin{tabular}{lrrrrr}
\toprule
\multicolumn{2}{c}{ } & \multicolumn{2}{c}{M3 (accelerating)} & \multicolumn{2}{c}{M0 (pure accumulator)} \\
\cmidrule(l{3pt}r{3pt}){3-4} \cmidrule(l{3pt}r{3pt}){5-6}
Word class & $n$ & slope & $R^2$ & slope & $R^2$\\
\midrule
nouns & 293 & -4.77 & 0.38 & -0.67 & 0.11\\
action words & 103 & -3.47 & 0.07 & -0.06 & 0.00\\
descriptive words & 62 & -1.68 & 0.03 & +0.49 & 0.03\\
function words & 90 & -3.18 & 0.13 & -0.22 & 0.01\\
\bottomrule
\end{tabular}
\endgroup{}

}

\end{table}%

\subsection{Matched Estimation of
Acceleration}\label{matched-estimation-of-acceleration}

One potential objection to the comparison of children and LMs in the
main text is that the estimators are different in two ways. First, the
estimator for children is Rasch-based, while the model from (\emph{34})
is a four-parameter logistic. Second, and perhaps more saliently, the
child estimator is for \(\kappa\) across children, while the LM
estimator is across words.

\begin{figure}[H]

\centering{

\includegraphics[width=1\linewidth,height=\textheight,keepaspectratio]{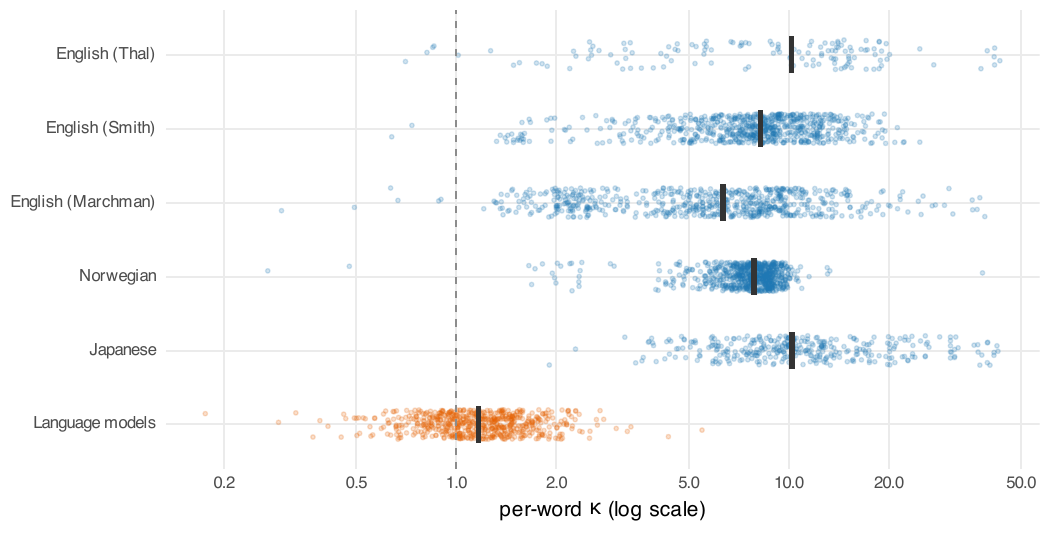}

}

\caption{\label{fig-perword-4pl}Per-word acceleration for children and
language models, both estimated with the same four-parameter logistic.
Points are words, with the median marked; language model values are
medians across the ten training seeds. The dashed line at one is the
pure-accumulator value. Child estimates are attenuated by pooling across
children and so understate the separation; the correction factors are
given in the table.}

\end{figure}

To address these issues, we fit the LM four-parameter estimator directly
to data on children's word acquisition, averaging across children to
produce a proportion of children producing each word at each age. (The
Thal and Japanese datasets had fewer children and less age diversity, so
not all words produced a viable curve estimate, but most words could be
estimated for the Smith, Marchman, and Norwegian datasets). Fig.
\ref{fig-perword-4pl} shows the results. Across all datasets, \(\kappa\)
estimates were robustly higher than LMs, though somewhat attenuated
relative to the M3 fits.

One possible source of this attenuation is the averaging of many
different acquisition curves across children. To estimate this
attenuation, we sampled from the fitted M3 parameters for \(\xi_i\),
\(\kappa_i\) and \(\delta_j\), and ran the identical pipeline over the
data to recover a correction factor (the ratio of the estimated by-word
\(\kappa\) to the population \(\kappa\) used to generate the data). We
then applied this correction factor to our empirical estimates. Table
\ref{tbl-perword-4pl} shows these corrected estimates, which match very
closely to the M3 values.

In sum, even with a completely matched estimator, acceleration diverges
substantially between children and LMs.

\begin{table}

\caption{\label{tbl-perword-4pl}Per-word acceleration estimated with the
language models' own estimator. For each word we fit the same
four-parameter logistic used on LM surprisal to the proportion of
children not yet producing that word, and take kappa = 0.434/scale.
Recovery is the fraction of the population kappa that the same pipeline
returns when run on data generated from the fitted M3, and is the factor
by which pooling across children attenuates the estimate; the corrected
estimate is the raw estimate divided by this factor. The final column is
the kappa M3 model reports for that dataset.}

\centering{

\centering\begingroup\fontsize{9}{11}\selectfont

\begin{tabular}{lrrrrr}
\toprule
Sample & words fitted & $\kappa$ raw [IQR] & recovery & $\kappa$ corrected & $\kappa$ (M3)\\
\midrule
English (Thal) & 135/681 & 10.18 [4.32, 14.84] & 98\% & 10.42 & 11.51\\
English (Smith) & 673/680 & 8.21 [6.04, 10.67] & 62\% & 13.21 & 12.95\\
English (Marchman) & 624/681 & 6.34 [3.12, 9.29] & 59\% & 10.66 & 10.65\\
Norwegian & 713/733 & 7.87 [6.97, 8.69] & 57\% & 13.81 & 13.25\\
Japanese & 343/447 & 10.22 [7.66, 15.56] & 86\% & 11.92 & 11.57\\
\addlinespace
Language models & 609/609 & 1.16 [0.93, 1.46] &  &  & \\
\bottomrule
\end{tabular}
\endgroup{}

}

\end{table}%

\subsection{Other Architectures}\label{other-architectures}

\begin{figure}[H]

\centering{

\includegraphics[width=1\linewidth,height=\textheight,keepaspectratio]{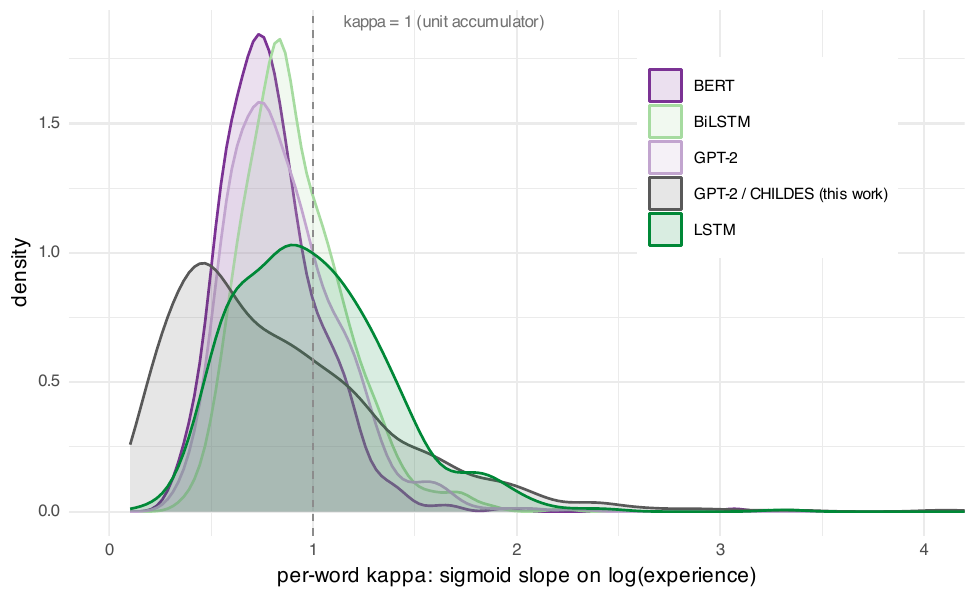}

}

\caption{\label{fig-si-cb-arch}Per-word acquisition slopes for the four
architectures studied in Chang \& Bergen (2022) -- BERT, GPT-2, BiLSTM,
LSTM -- trained on BookCorpus + WikiText-103. CHILDES-trained GPT-2
models (training axis) are overlaid in gray.}

\end{figure}

The acceleration results we report in the main text are not specific to
our choice of the GPT-2 architecture. We re-analyze the published fits
of (\emph{34}). Fig. \ref{fig-si-cb-arch} shows a similar distribution
of per-word \(\kappa\) values for the four different architectures
reported in that study, which were trained on standard LM corpora rather
than child-directed input.

\end{document}